\documentclass[runningheads]{llncs}
\usepackage{amsmath}
\usepackage{verbatim}
\usepackage[T1]{fontenc}
\usepackage{graphicx}
\begin{document}
\title{Optimizing Prosthetic Wrist Movement: A Model Predictive Control Approach}
\titlerunning{ Model Predictive Control}
%
\author{ Francesco Schetter $^{1}$ \and Shifa Sulaiman*$^{1}$ \orcidID{0000-0002-5330-0053} \and
Shoby George$^{2}$  \and
Paolino De Risi$^{1}$  \and
Fanny Ficuciello$^{1}$ } 
\authorrunning{S. Francesco \textit{et al.}}
%
\institute{$^{1}$ Department of Information Technology and Electrical Engineering, Università degli Studi di Napoli Federico II, Claudio, 21, 80125 Napoli, Italy \\
$^{2}$ Genrobotic Innovations Pvt. Ltd., Kerala, India
\email{*ssajmech@gmail.com}}
\maketitle              
\begin{abstract}


The integration of advanced control strategies into prosthetic hands is essential to improve their adaptability and performance. In this study, we present an implementation of a Model Predictive Control (MPC) strategy to regulate the motions of a soft continuum wrist section attached to a tendon-driven prosthetic hand with less computational effort. MPC plays a crucial role in enhancing the functionality and responsiveness of prosthetic hands. By leveraging predictive modeling, this approach enables precise movement adjustments while accounting for dynamic user interactions. This advanced control strategy allows for the anticipation of future movements and adjustments based on the current state of the prosthetic device and the user’s intentions. Kinematic and dynamic modelings are performed using Euler‐Bernoulli beam and Lagrange's methods respectively. Through simulation and experimental validations, we demonstrate the effectiveness of MPC in optimizing wrist articulation and user control. Our findings suggest that this technique significantly improves the prosthetic hand’s dexterity, making movements more natural and intuitive. This research contributes to the field of robotics and biomedical engineering by offering a promising direction for intelligent prosthetic systems.

\end{abstract}
\section{INTRODUCTION}
Soft robotic prostheses \cite{ref1} represent a significant breakthrough, offering individuals with limb disabilities a more comfortable and natural range of motions compared to conventional rigid prosthetics. The incorporation of soft continuum components enables intricate movements, making these devices suitable for a wide array of applications. Additionally, elastic wires integrated into these soft sections function like tendons, providing flexibility, lightweight characteristics, cost-effectiveness, and the ability to endure substantial tensile forces \cite{wmrac}.

Model predictive control (MPC) is essential in the development of prosthetic hands as it enables a more intuitive and seamless interaction between the user and the device. By forecasting the necessary actions and adjusting the control inputs accordingly, this approach ensures that the prosthetic hand can execute complex movements with precision. This capability is particularly important for users who require fine motor skills for daily activities, as it allows for smoother transitions and greater adaptability to different scenarios.

Furthermore, the integration of MPC strategy in prosthetic hands contributes to the overall safety and reliability of these devices. By continuously monitoring the system's performance and making real-time adjustments, potential issues can be identified and addressed before they lead to malfunctions or accidents. 
Major contributions of this work are as follows:

\begin{itemize}
    \item Kinematic and dynamic modelings of a soft continuum wrist using Euler‐Bernoulli beam and Lagrange's method respectively.
    \item Development of an MPC scheme for the wrist motions.
    \item Simulation studies to demonstrate the advantages of the proposed controller.
    \item Experimental validations proving the effectiveness of the proposed controller during real-time implementations with reduced computational effort. 
\end{itemize}

The implementation of MPC plays a crucial role in the domain of prosthetic hand motion management. This significance arises from its ability to optimize movements and functionalities of prosthetic devices, ensuring that they respond accurately to the user's intentions and environmental conditions. By utilizing predictive algorithms, this control method can anticipate future states and adjust the prosthetic's actions accordingly, leading to more natural and efficient hand movements. Moreover, MPC enhances the adaptability of prosthetic hands by allowing for real-time adjustments based on sensory feedback. This capability is essential for users who require precise control over their prosthetic devices, as it enables the hands to perform complex tasks with greater ease and reliability. The integration of such advanced control strategies not only improves the overall user experience but also contributes to the development of more sophisticated and responsive prosthetic technologies. In addition, the relevance of MPC extends beyond mere motion management; it also encompasses the potential for learning and improvement over time. As users interact with their prosthetic hands, the control system can gather data and refine its predictive models, leading to enhanced performance tailored to individual preferences and needs. This continuous learning process is vital for the evolution of prosthetic technology, ultimately aiming to provide users with a level of functionality that closely resembles that of a natural hand.

Spinelli \textit{et al.} \cite{mpc1} proposed a modular MPC framework for soft continuum manipulators, integrating internal and external constraints. The approach successfully implemented Task-Space MPC, improving dynamic control. The method relies on Piece-wise Constant Curvature (PCC) assumptions, which may limit its applicability to highly deformable soft robots. Johnson \textit{et al.} \cite{mpc2} introduced a hybrid modeling approach, combining machine learning with first-principles models to enhance MPC performance. The method improved control accuracy by 52\% on average. The reliance on large datasets for training deep learning models may hinder real-time adaptability. Pal \textit{et al.} \cite{mpc3} proposed a data-driven MPC design using Bayesian optimization for cable-actuated soft robots. Instead of modeling complex dynamics, the approach searches for an optimal low-dimensional prediction model iteratively. The method required multiple iterations to converge, which may limit real-time applications.

Yang \textit{et al.} \cite{mpc4} introduced a distributionally robust MPC (DRMPC) scheme based on neural network modeling to achieve trajectory tracking control for robotic manipulators with state and control torque constraints. Their approach converted motion data into a linear prediction model and applies chance constraints to optimize control decisions. The reliance on statistical analysis of modeling errors may introduce computational complexity, limiting real-time applications. Huang \textit{et al.} \cite{mpc5} developed a physics-learning hybrid modeling approach combining absolute nodal coordinate formulation (ANCF) with multilayer neural networks (MLNN) to enhance dynamic control accuracy. The method significantly reduced tracking errors and improves real-time simulation efficiency. The complexity of hybrid modeling required extensive parameter tuning, making implementation challenging for new robotic designs.

Gonzales \textit{et al.} \cite{mpc6} introduced a multi-agent receding-horizon feedback motion planning approach using Probably Approximately Correct Nonlinear MPC (PAC-NMPC). The method enhanced formation control and obstacle avoidance in dynamic environments while accounting for model and measurement uncertainty. The computational complexity of PAC-NMPC limited scalability for large multi-robot teams. Kalibala \textit{et al.} \cite{mpc7} developed a deep neural network (DNN)-based MPC framework for pressure-driven vine robots. Their approach significantly improves control performance and computational efficiency, reducing computation time by a factor of 11 compared to traditional nonlinear first-principles models. The data-driven nature of the model requires extensive training, which may limit adaptability to new robotic designs. Chen \textit{et al.} \cite{mpc8} proposed Vision-Language Model Predictive Control (VLMPC), integrating vision-language models (VLMs) with MPC for robotic manipulation planning. The approach enhances environmental perception and improves trajectory generation accuracy. The computational overhead of integrating vision-language models can restrict real-time applications.

An extensive examination of the literature on soft continuum robots revealed that soft robots face multiple challenges, including inadequate kinematic and dynamic modeling methods, suboptimal control strategies, and increased computational requirements. This research introduces modeling techniques and control strategies designed to develop an MPC for a soft wrist component incorporated into a prosthetic hand, with the goal of enhancing response times and reducing computational demands.
The organization of this paper is as follows:  Section 2 provides an overview of the modeling methodologies employed. The MPC strategy is presented in Section 3. Section 4 highlights the results from simulations and experiments. Lastly, Section 5 offers a conclusion to the study.

\section{Mathematical model of a soft wrist section}
\begin{figure}[hbt!]
\centerline{\includegraphics[width=0.6\textwidth]{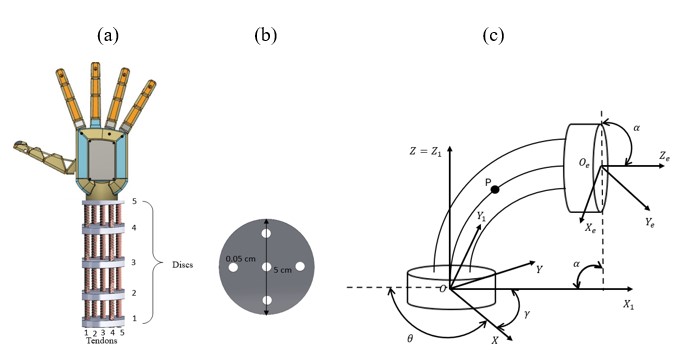}}
\caption{Soft wrist section (a) Conceptual design of wrist section attached to hand (b) Dimension of disc (c) Bending structure of wrist section}
\label{fig.1}
\end{figure}

\begin{figure}[hbt!]
\centerline{\includegraphics[width=0.3\textwidth]{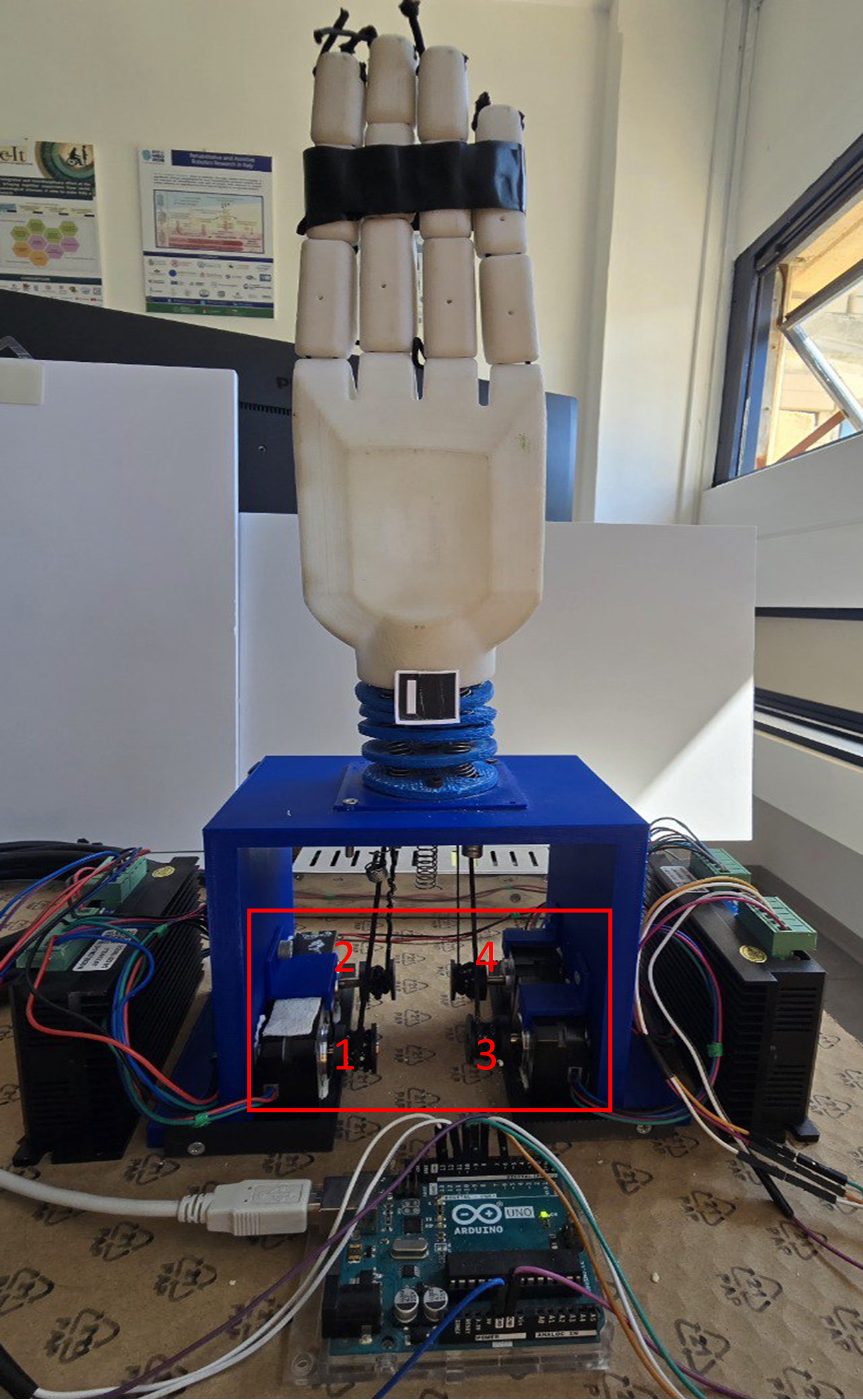}}
\caption{Fabricated model}
\label{fabricated}
\end{figure}

The proposed design for the soft wrist segment, as outlined in \cite{ref15}, consists of five rigid discs, five springs, and five flexible tendons attached to a prosthetic hand named 'PRISMA HAND II' \cite{prisma}, as shown in Fig. \ref{fig.1}(a). The dimensions of the rigid discs incorporated in this wrist segment are presented in Fig. \ref{fig.1}(b). Furthermore, Fig. \ref{fig.1}(c) illustrates the bending configuration of the soft wrist segment with length $l$, radius $r$, and a bending angle of $\alpha$.  
The rigid discs incorporate the springs and tendons, which are attached to a stable platform. By exerting precise tensions on each tendon via a motor, the intended movements of the wrist segment can be achieved. Four peripheral tendons are incorporated to enable rotational movements in four different directions. Tendons one and two are engaged for radial deviation of the wrist, while tendons four and five facilitated movements in the ulnar direction. Additionally, tendons one and four are responsible for extension motions, whereas flexion is managed by tendons two and five. The lowest disc (disc one) was secured to a stable platform, and the highest disc (disc five) was connected to the hand. Even though the wrist section is actuated using four tendons, there is also a middle tendon used as a primary backbone to provide stability during motions.  Fig. \ref{fabricated} illustrates the constructed model of the wrist section combined with a prosthetic hand.
Variations in tendon tensions result in distinct bending moments on the soft wrist segment, enabling its behavior to be represented as a cantilever beam under bending stress. The placement of the end effector relative to the wrist's curvature is established based on bending beam theory, as indicated in \cite{bending}. \\ \\
The transformation matrix, $T$ of the last disc five  with respect to the base disc one is expressed as follows:
\begin{equation}
    T = 
    \begin{bmatrix}
        R & P \\ 0 & 1
    \end{bmatrix}
\end{equation}
where rotation matrix, $R$ is obtained as given in Eq. \eqref{rotation}
\begin{equation}
    R=Rot(Z,\gamma)Rot(Y,\alpha)Rot(Z,-\gamma)=
    \begin{bmatrix}
        c^2\gamma c\alpha+s^2\gamma & c\gamma s\gamma c\alpha-c\gamma s\gamma & c\gamma s\alpha \\
        c\gamma s\gamma c\alpha-c\gamma s\gamma & s^2\gamma c\alpha+c^2\gamma & s\gamma s\alpha \\
        -c\gamma s\alpha & -s\gamma c\alpha & c\alpha
        
    \end{bmatrix}
     \label{rotation}
\end{equation}
Translational matrix $P$ is given in Eq. \eqref{point}.
\begin{equation}
    P=[
    \begin{array}{ccc}
       x & y & z  
    \end{array}
    ]^T= 
    \begin{bmatrix}
        \frac{l}{\alpha}(1-\cos{\frac{s \alpha}{l}})\cos{\gamma} &
        \frac{l}{\alpha}(1-\cos{\frac{s \alpha}{l}})\sin{\gamma} &
        \frac{l}{\alpha}\sin{\frac{s \alpha}{l}}
    \end{bmatrix}
    \label{point}
\end{equation}
The kinetic energy of the wrist section motions is obtained by calculating the derivative of the positions given in Eq. \eqref{point}. The
velocities of motion can be expressed as follows:
  
\begin{equation}
    \begin{cases}
        \begin{split}
           \frac{dx}{dt} &=\frac{1}{\alpha}[s\sin{\frac{s\alpha}{l}}\cos{\gamma}-\frac{l}{\alpha}(1-\cos{\frac{s\alpha}{l}})\cos{\gamma}]\frac{d\alpha}{dt} \\
           & -\frac{1}{\alpha}(1-\cos{\frac{s\alpha}{l}})\sin{\gamma}\frac{d\gamma}{dt}
        \end{split}
         \\
         \begin{split}
             \frac{dy}{dt} &=\frac{1}{\alpha}[s\sin{\frac{s\alpha}{l}}\sin{\gamma}-\frac{l}{\alpha}(1-\cos{\frac{s\alpha}{l}})\sin{\gamma}]\frac{d\alpha}{dt} \\
           & +\frac{1}{\alpha}(1-\cos{\frac{s\alpha}{l}})\cos{\gamma}\frac{d\gamma}{dt}
         \end{split}
         \\
         \begin{split}
             \frac{dz}{dt} &=\frac{1}{\alpha}(s\cos{\frac{s\alpha}{l}}-\frac{l}{\alpha}\sin{\frac{s\alpha}{l}})\frac{d\alpha}{dt} 
         \end{split}
    \end{cases}
    \label{12}
\end{equation}
The kinetic energy of the primary backbone (central tendon), $E_{k1}$ of the soft wrist can be obtained as follows:

\begin{equation}
    E_{k1}=\frac{1}{2}\int_0^l
    \begin{bmatrix}
        (\frac{dx}{dt})^2+(\frac{dy}{dt})^2+(\frac{dz}{dt})^2
    \end{bmatrix}
    \rho Ads
    \label{13}
\end{equation}
where $\rho$ and$A$ represent density and cross‐sectional area of the wrist section. Substituting Eq. \eqref{12} in Eq. \eqref{13}, we obtained the kinetic energy  as given in Eq. \eqref{16}.
\begin{equation}
    E_{k1}=\frac{1}{6}m_1l^2(\frac{d\alpha}{dt})^2K_1+\frac{1}{8}m_1l^2(\frac{d\gamma}{dt})^2K_2 
    \label{16}
\end{equation}  
where $m_1$ is the mass of the primary backbone and $K_1$ and
$K_2$ are the kinetic energy equivalent factors. Kinetic energy coefficients, $K_1$ and $K_2$ are determined as given in Eqs. \eqref{new1} and \eqref{new2}.
\begin{equation}
 K_1=(\alpha^3+6\alpha-12\sin{\alpha}+6\alpha \cos{\alpha})/\alpha^5  
     \label{new1}
\end{equation}
\begin{equation}
    K_2=(6\alpha 8\sin{\alpha}+\sin{2\alpha})/\alpha^3 \label{new2}
\end{equation}
From Eq. \eqref{new1} and \eqref{new2}, we can express $K_1$ and $K_2$ as a function of bending angle $\alpha$ . The two
Eqs. can be simplified using least square fit, as given in Eqs. \eqref{equ1} and \eqref{sim2}:
\begin{equation}
    K_1=-0.00426\alpha^2-0.00277\alpha+0.15085
    \label{equ1}
\end{equation}
  
\begin{equation}  K_2=-0.05567\alpha^3+0.2328\alpha^2+0.006216\alpha-0.00406
    \label{sim2}
\end{equation}
The transformation between cartesian and joint spaces can be expressed as follows:

\begin{equation}
    \begin{cases}
        q_1=r\alpha \cos(\gamma) \\
        q_2=r\alpha \cos(-\gamma+\theta) \\
        q_3=r\alpha \cos(\gamma+\theta)
    \end{cases}
    \label{11}
\end{equation}
where, $q_i,(i=1, 2, 3)$ is the length of each driving wire and $r$ is the distance from each secondary backbone to the
primary backbone (the secondary backbone tendons are
equidistant from the primary backbone), and $\theta=2\pi/3$. The driving velocities are obtained by performing derivatives of Eq. \eqref{11}, and obtained as follows:
\begin{equation}
    \begin{cases}
        \frac{dq_1}{dt}=r\cos(\gamma)\frac{d\alpha}{dt}-r\alpha\sin(\gamma)\frac{d\gamma}{dt} \\
        \frac{dq_2}{dt}=r\cos(-\gamma+\theta)\frac{d\alpha}{dt}+r\alpha\sin(-\gamma+\theta)\frac{d\gamma}{dt} \\
        \frac{dq_3}{dt}=r\cos(\gamma+\theta)\frac{d\alpha}{dt}-r\alpha\sin(\gamma+\theta)\frac{d\gamma}{dt}
    \end{cases}
    \label{21}
\end{equation}
The secondary backbone consisted of four tendons. However, the tendons present on each side during motion are considered as a single tendon for the analysis. For example, if the tendons are rotating in ulnar deviation direction, tendons four and five are considered as a single tendon and tendons one and two are considered as two tendons itself. 
The total kinetic energy, $E_{k2}$ of the secondary backbone consisted of four tendons are given in equation \eqref{KE}

\begin{equation}
 E_{k2} = E_{k11} + E_{k22}
  \label{KE}
\end{equation}
where $E_{k11}$ = $E_{k1}$ and the second component,  $E_{k22}$ arises from the driven kinetic energy as given in  following Eq.:
\begin{equation}
    E_{k22}=\frac{1}{2}m_1
    \begin{bmatrix}
        (\frac{dq_1}{dt})^2+(\frac{dq_2}{dt})^2+(\frac{dq_3}{dt})^2
    \end{bmatrix}
    \label{22}
\end{equation}

Substituting Eq. \eqref{21} in Eq. \eqref{22}, we obtain the following Eqs. \eqref{eq1} - \eqref{26}

\begin{gather}
    E_{k2}=\frac{1}{2}m_2 \Bigl [(\frac{d\alpha}{dt})^2K_3+\frac{d\alpha}{dt}\frac{d\gamma}{dt}K_4+(\frac{d\gamma}{dt})^2K_5 \Bigr ] 
    \label{eq1} \\
    K_3=r^2
    \begin{bmatrix}
        \cos^2(\gamma)+\cos^2(-\gamma+\theta)+\cos^2(\gamma+\theta)
    \end{bmatrix}
    \label{24}\\
    K_4=r^2\alpha
    \begin{bmatrix}
        -\sin(2\gamma)+\sin(2(-\gamma+\theta))-\sin(2(\gamma+\theta))
    \end{bmatrix}
    \label{25}\\
    K_5=r^2\alpha^2
    \begin{bmatrix}
        \sin^2(\gamma)+\sin^2(-\gamma+\theta)+\sin^2(\gamma+\theta)
    \end{bmatrix}
    \label{26}
\end{gather}
where $m_2$ is the mass of the secondary backbone and $K_3$,
$K_4$ and $K_5$ are kinetic energy equivalent factors. The kinetic energy of the  discs
can be obtained as:

\begin{equation}
    E_{k3}=\frac{1}{2}m_3(\frac{d\alpha}{dt})^2K_6+\frac{1}{2}m_3(\frac{d\gamma}{dt})^2K_7
    \label{30}
\end{equation}
where, $m_3$ is the mass of a disk and $K_6$ and $K_7$ are kinetic energy equivalent factors.
If $n$ and $h$ are determined, the
kinetic energy equivalent factors $K_6$ and $K_7$ can be
expressed as a function of bending angle $\alpha$ . Assuming
that $n=5$ and $h=15~mm$, the form of $K_6$ and $K_7$ can be
simplified by using least square fit, shown as follows:

\begin{equation}
    K_6 = (-0.00043\alpha^2-0.00031\alpha+0.01435)/2
    \label{33}
\end{equation}
\begin{equation}
    K_7=(-0.00394\alpha^3+0.01575\alpha^2+0.00131\alpha-0.00047)/2
    \label{34}
\end{equation}
For a continuum robot, the total potential energy is comprised of two components: elastic potential energy and gravitational potential energy. In this context, the gravitational potential energy can be considered negligible compared to the elastic potential energy. The elastic energy, $E_p$ associated with the wrist section with Young's Modulus $E$ and inertia $I$  is given as follows:
\begin{equation}
    E_p=\frac{2EI}{l}\alpha^2
\end{equation}
The Lagrange Eq. of the wrist section is expressed as follows:
\begin{equation}
    \frac{d}{dt}\frac{\partial E_k}{\partial \Dot{p_j}}-\frac{\partial E_k}{\partial p_j}+\frac{\partial E_p}{\partial p_j}=Q_j,(j=1,2)
\end{equation}
where $Q_j$ represents the generalized force of system, $E_k=E_{k1}+E_{k2}+E_{k3}$, $p_1=\alpha$ and $p_2=\gamma$.
The dynamical Eq. of the wrist is obtained as follows:

\begin{equation}
    \begin{bmatrix}
        M_{11} & M_{12}\\
        M_{21} & M_{22}
    \end{bmatrix}
    \begin{bmatrix}
        \Ddot{\alpha}\\ \Ddot{\gamma}
    \end{bmatrix}
    +
    \begin{bmatrix}
        C_{11} & C_{12} & C_{13}\\
        C_{21} & C_{22} & C_{23}
    \end{bmatrix}
    \begin{bmatrix}
        \Dot{\alpha}^2\\ \Dot{\alpha}\Dot{\gamma}\\ \Dot{\gamma}^2
    \end{bmatrix}
    +
    \begin{bmatrix}
        K_{11} & K_{12}\\
        K_{21} & K_{22}
    \end{bmatrix}
    \begin{bmatrix}
        \alpha \\ \gamma
    \end{bmatrix}
    =
    \begin{bmatrix}
        D_{11} & D_{12}\\
        D_{21} & D_{22}
    \end{bmatrix}
    \begin{bmatrix}
        F_1 \\ F_2
    \end{bmatrix}
\end{equation}
where $M_{ij}$,$C_{ij}$,$K_{ij}$,$D_{ij}$ are moment of inertia, Coriolis, stiffness, actuation matrix elements respect to each rotation angle.

In the context of planar motion, where $\gamma = 0$, we determined the Eq. of motion as follows:

\begin{equation}
    M(\alpha)\Ddot{\alpha}+C(\alpha)\Dot{\alpha}^2+K\alpha=DF
    \label{40}
\end{equation}

where:$$D=r\cos(\gamma)$$ $$K=\frac{4EI}{l}$$ $$C=-\frac{1}{6}(4m_2l^2\frac{\partial K_1}{\partial \alpha})+3m_2(\frac{\partial K_3}{\partial \alpha})+3m_3(\frac{\partial K_6}{\partial \alpha})$$ $$M=\frac{1}{3}(4m_2l^2K_1+3m_2K_3+3m_3K_6)$$

\section{ Inverse Dynamic Model Predictive Control developed for the soft wrist section }

\begin{figure}[htb!]
    \centering
    \includegraphics[width=0.8\linewidth]{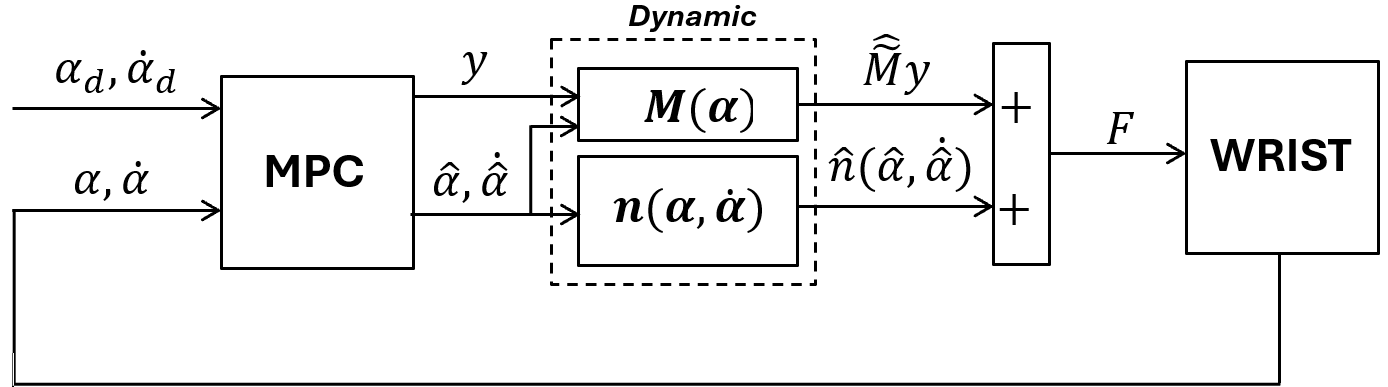}
        \caption{Inverse Dynamic Model Predictive Control scheme}
        \label{control_scheme}
\end{figure}

The advancement of prosthetic technology has increasingly focused on achieving natural and adaptive movement to enhance user experience and functionality. A crucial component of the method is the implementation MPC, particularly Inverse Dynamic Model Predictive Control, for managing the movements of soft wrist components in prosthetic hands. By utilizing predictive modeling, this approach enables precise, real-time adjustments to control inputs, significantly improving motion fluidity and responsiveness. Unlike traditional control methods, Inverse Dynamic MPC anticipates future states, allowing for smoother, more intuitive interactions that reduce cognitive load and enhance usability. The control strategy developed for the wrist section is shown in Fig. \ref{control_scheme}. Desired states such as bending angles ($\alpha$) and rate of bending angles ($\dot \alpha$) are fed to the MPC block. Optimised output $y$ and the predicted state values $\hat{\alpha}$ and $\Dot{\hat{\alpha}}$ from the Model Predictive Control (MPC) are utilized to calculate the control law, $u=F$, and estimated $\hat{\Tilde{M}}$ and $\hat{n}(\hat{\alpha},\Dot{\hat{\alpha}})$.

We can rewrite dynamic equation of wrist section given in eq. \eqref{40} as follows:
\begin{equation}
    \Tilde{M}\Ddot{\alpha}+\Tilde{C}(\alpha, \Dot{\alpha})\Dot{\alpha}+\Tilde{K}\alpha=u
\end{equation}
where $\Tilde{M}(\alpha)=D^{-1}M(\alpha)$, $\Tilde{C}(\alpha, \Dot{\alpha})=D^{-1}C(\alpha)\Dot{\alpha}$,  $\Tilde{K}=D^{-1}K$. $u$ is chosen as given in equation \eqref{f} for a feedback linearization control action.
\begin{equation}
    u = \Tilde{M}y + n(\alpha, \Dot{\alpha})
     \label{f}
\end{equation}
where $n(\alpha, \Dot{\alpha})=\Tilde{C}(\alpha, \Dot{\alpha})\Dot{\alpha}+\Tilde{K}\alpha$, and we can obtain optimised output, $y$ as follows:
\begin{equation}
    \Ddot{\alpha}=y
    \label{inverse}
\end{equation}
From Eq. \eqref{inverse}, we can compute the state space form as follows:
\begin{equation}
    A =
    \begin{bmatrix}
        0 & 0 \\ 1 & 0
    \end{bmatrix},
    B = 
    \begin{bmatrix}
        1 \\ 0
    \end{bmatrix},
    C = 
    \begin{bmatrix}
        1 & 0 \\ 0 & 1
    \end{bmatrix},
    D = 0
\end{equation}
The prediction and control horizons are established as $p=10$ and $\nu=5$. The cost function that the MPC aims to optimize is as follows:
\begin{equation}
    J(z_k)=J_x(z_k)+J_{\Delta u}(z_k)+J_\varepsilon(z_k)
\end{equation}
where,
\begin{gather}
    J_x(z_k) = \sum_{j=1}^{n_x}\sum_{i=1}^p \Bigl \{ \frac{w_{i,j}^x}{s_j^x}[r_j(k+i|k)-x_j(k+i|k)] \Bigr \}^2 \\
    J_{\Delta u}(z_k) = \sum_{j=1}^{n_u}\sum_{i=0}^{p-1} \Bigl \{ \frac{w_{i,j}^{\Delta u}}{s_j^u}[u_j(k+i|k)-u_j(k+i-1|k)] \Bigr \}^2 \\
    J_\varepsilon(z_k) = \rho_\varepsilon \varepsilon _k2
\end{gather}
where,
\begin{itemize}
    \item[$\varepsilon_k$] Slack variable at the control interval $k$
    \item[$\rho_\varepsilon$] Constraint violation penalty weight
    \item[$z_k$] QP decision variables vector
    \item[$k$] Current control interval
    \item[$p$]  Prediction horizon
    \item[$n_x$] Number of plant output variables
    \item[$n_u$] Number of manipulated variables
    \item[$s_j^x$] Scale factor for the $j$th plant output
    \item[$s_j^u$] Scale factor for the $j$th MV 
    \item[$w_{i,j}^x$] Tuning weight for the jth plant output at the $i$th prediction horizon step
    \item[$w_{i,j}^{\Delta u}$] Tuning weight for the $j$th MV movement at the $i$th prediction horizon step 
    \item[$r_j(k+i|k)$]  Reference value for the jth plant output at the $i$th prediction horizon step
    \item[$x(k+i|k)$] Predicted value of the jth plant output at the $i$th prediction horizon step 
\end{itemize}

The KWIK active-set algorithm is employed to address the quadratic programming (QP) problem. The active constraints include a position constraint that limits motion to a maximum of $\pi / 4$ radians and a rate of change for the manipulated variable to ensure a smoother signal. Additionally, the predicted state values $\hat{\alpha}$ and $\Dot{\hat{\alpha}}$ from the Model Predictive Control (MPC) are utilized to calculate an estimated $\hat{n}(\hat{\alpha},\Dot{\hat{\alpha}})$ and $\hat{\Tilde{M}}$.

\section{Results and Discussions}
This research introduces an MPC tailored for the soft wrist of a prosthetic hand. The kinematic and dynamic modeling of the wrist are performed using Timoshenko beam theory. To evaluate the effectiveness of the proposed position controller while the soft wrist section is in motion with a payload, a series of simulations and experimental tests were conducted.

\subsection{Simulation study}

\begin{figure}[hbt!]
    \centering
\includegraphics[width=0.6\textwidth]{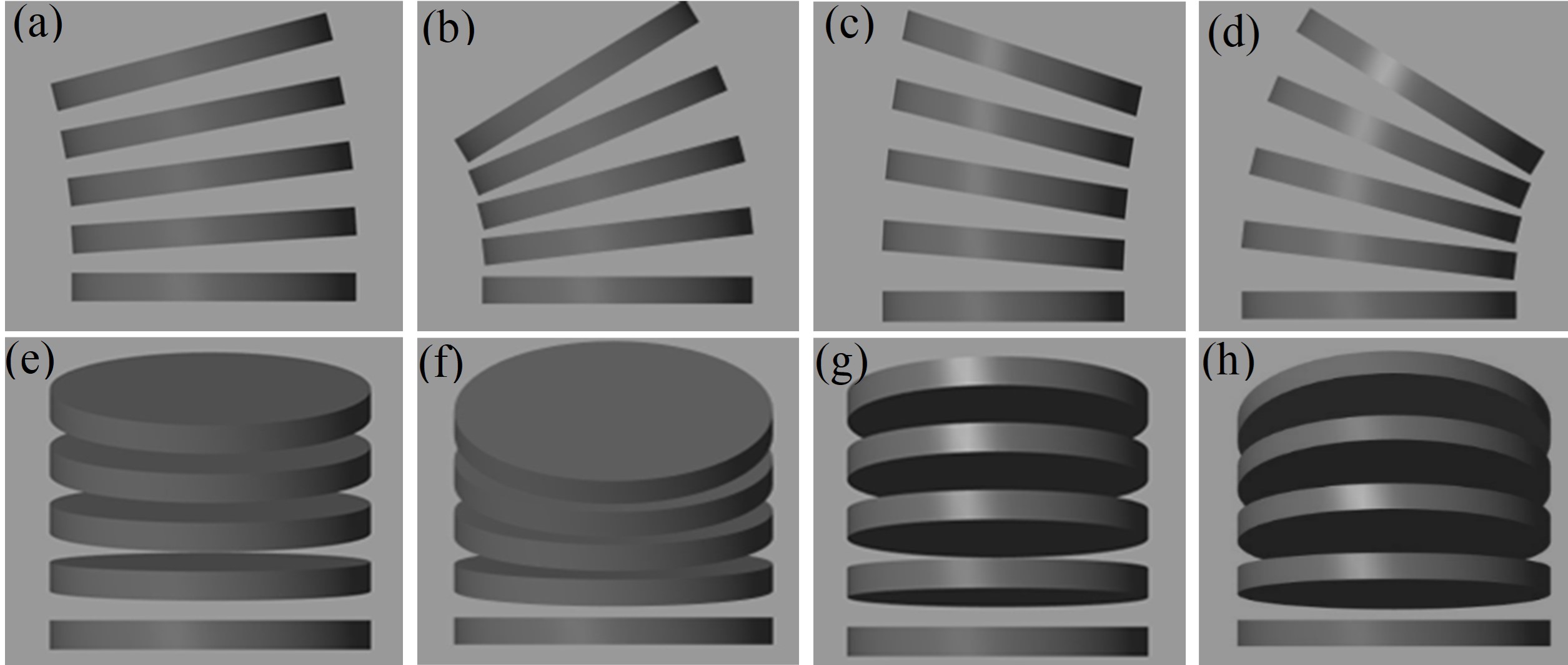}
   \caption{Motion of wrist (a) Radial-1 (b) Radial-2 (c) Ulnar-1 (d) Ulnar-2 (e) Flexion-1 (f) Flexion-2 (g) Extension-1 (h) Extension-2}
    \label{Motion of wrist}
\end{figure}

\begin{figure*}[hbt!]  
    \includegraphics[width=0.5\textwidth,height =1.92 in]{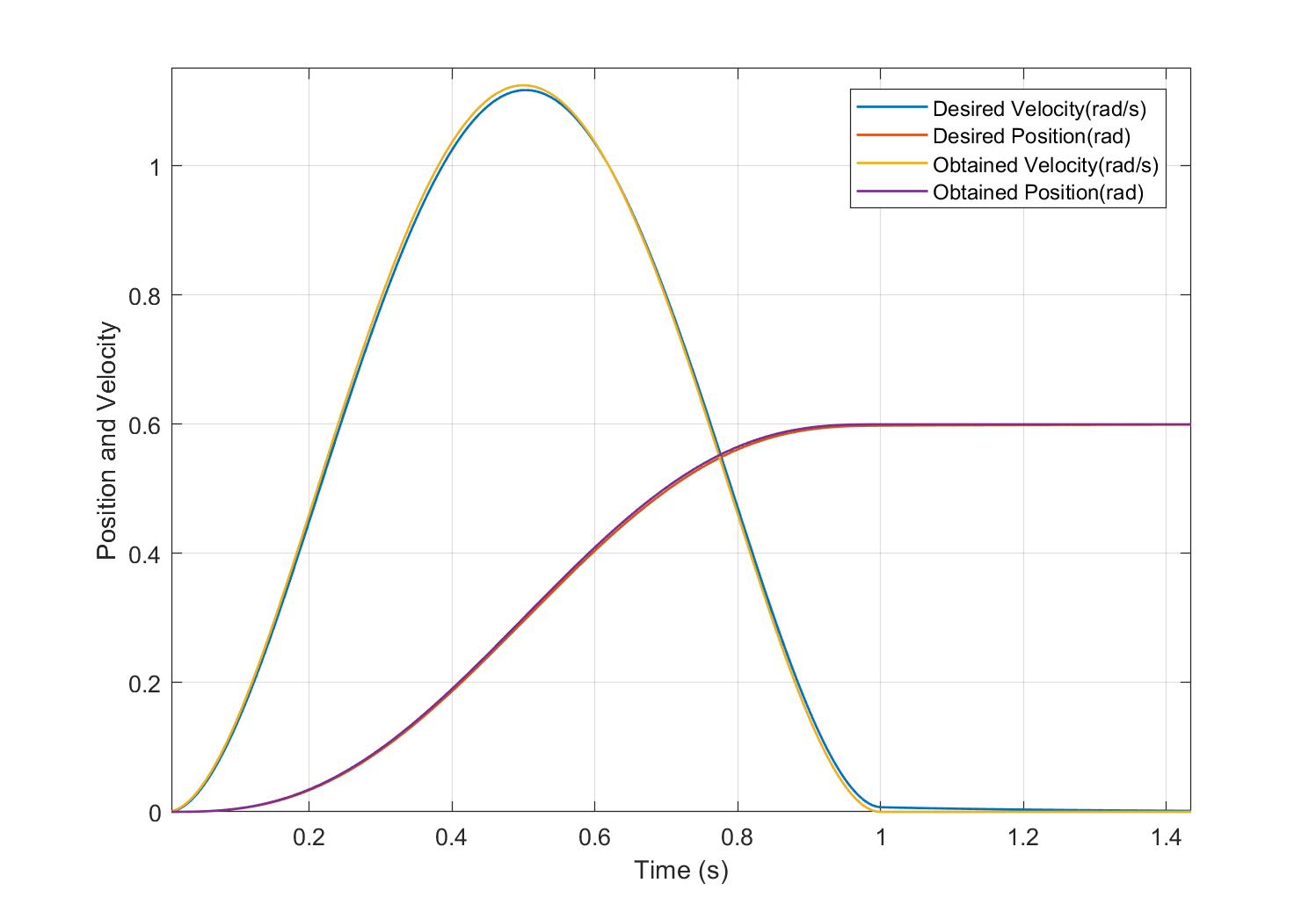}
     \includegraphics[width=0.5\textwidth,height =1.9 in]{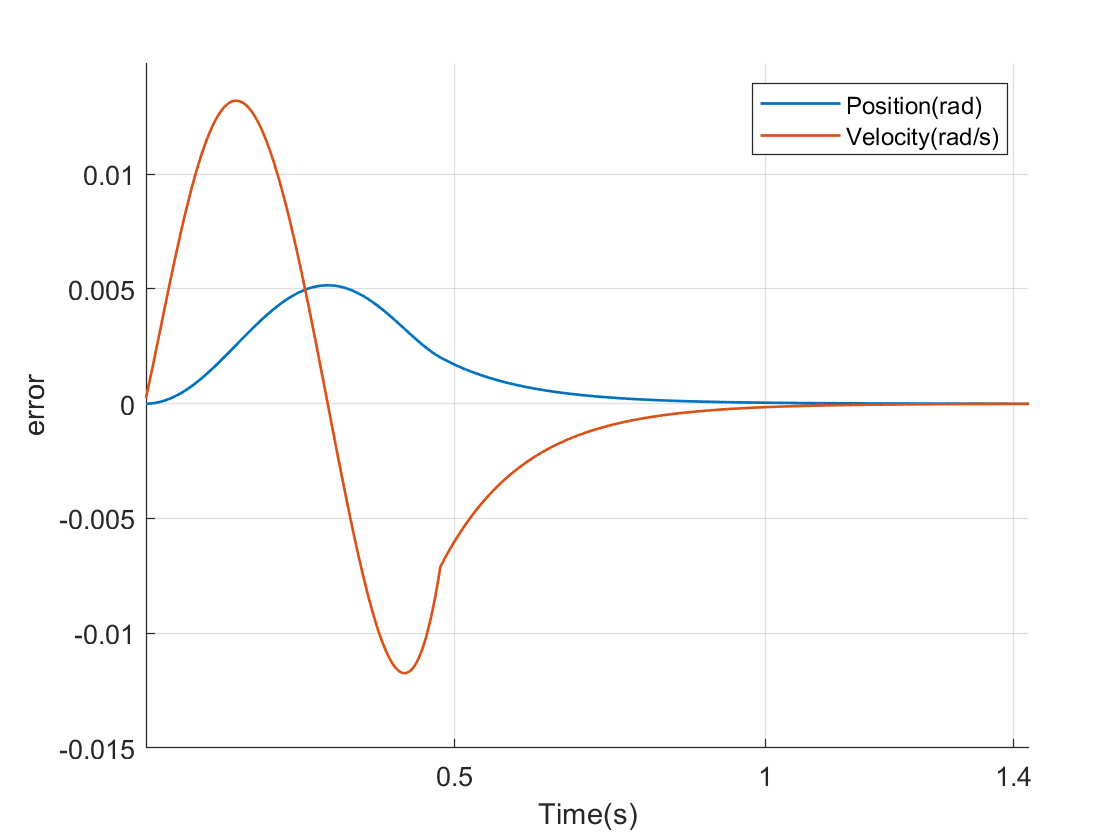}
    \caption{(a)Comparison of response with reference signal (b)Error in motions during simulation}
   \label{response}
\end{figure*}

The control scheme simulation is conducted using Simulink, a software platform based on MATLAB, on a PC equipped with an Intel Core Ultra 7 processor and 16 GB of RAM. The wrist segment is capable of moving along trajectories in the directions of radial deviation, ulnar deviation, flexion, and extension, as illustrated in Figs. \ref{Motion of wrist} (a) - (h).
The wrist segment is observed to flex from its initial position, as depicted in Fig. \ref{Motion of wrist}, achieving a final bending angle of $35^{0}$ in the direction of ulnar deviation relative to disc 5, which is attached to the hand. Desired and obtained positions and velocities are shown in Fig. \ref{response}(a). Additionally, the discrepancies in positions and velocities recorded during the simulation are presented in Fig. \ref{response}(b).
The values for Root Mean Square (RMSE), settling time, and steady state error are determined as $2.1\times 10^{-3}~rad$, $1.2~s$, and $0.4 \times 10^{-5}~rad$, respectively. 


\subsection{Experimental validation}
\begin{figure}[hbt!]
    \centering
  \includegraphics[width=0.7\textwidth,height =1.8 in]{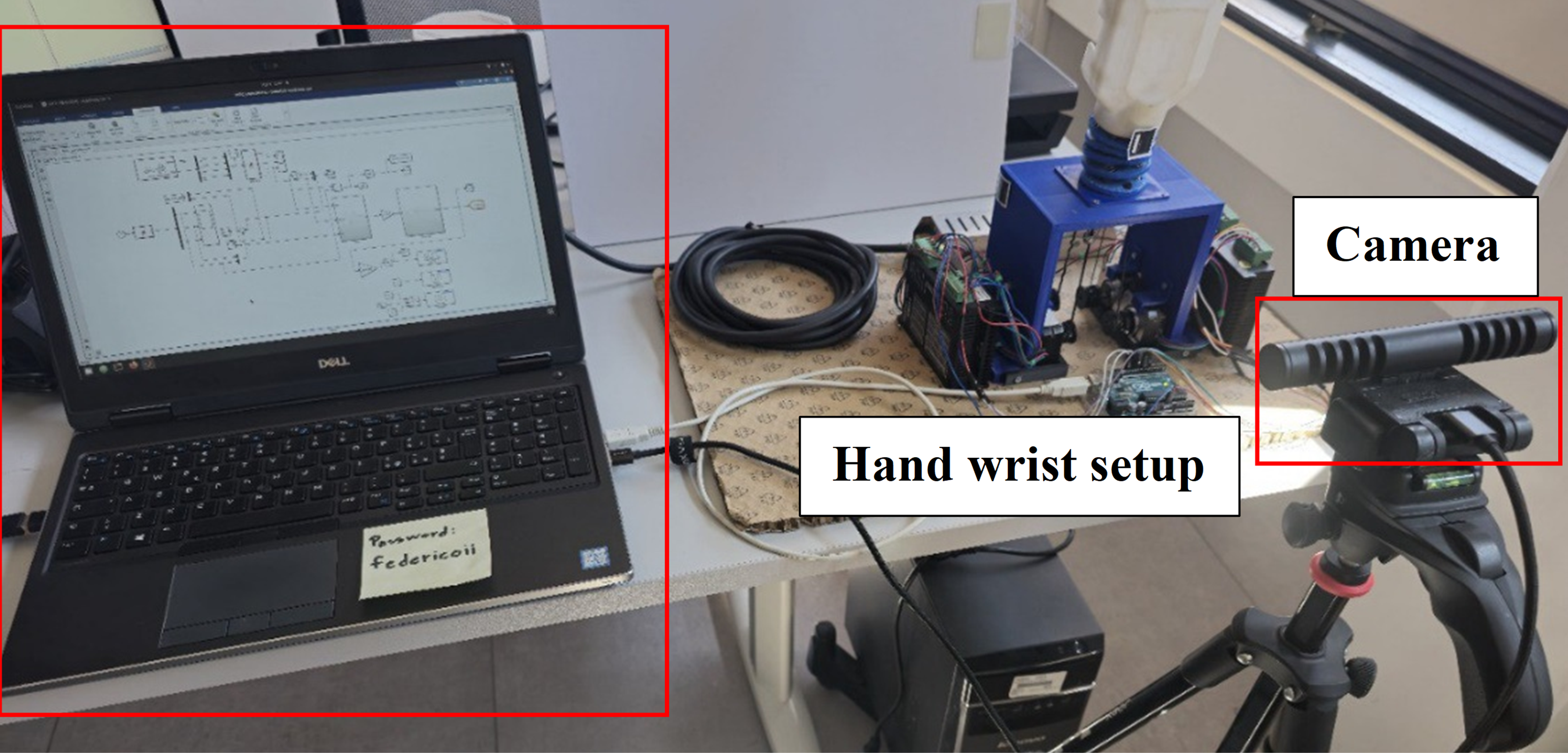}
    \caption{Experimentation set up}
   \label{exp_setup}
\end{figure}

\begin{figure}[hbt!]
    \centering
\includegraphics[width=0.55\textwidth, height = 3.5 in]{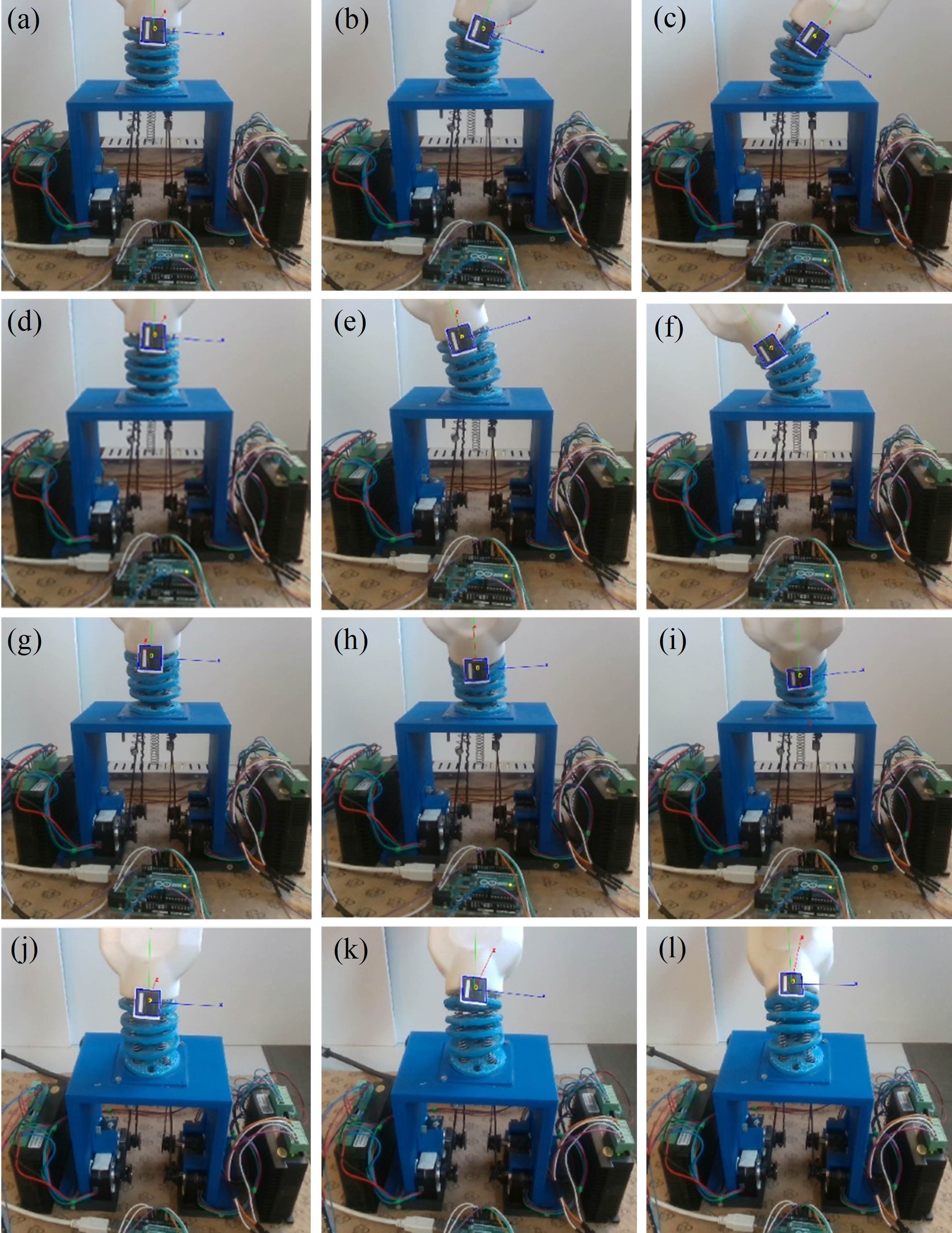} 
    \caption{Motion of the wrist section along with hand (a)-(c)Ulnar (d)-(f)Radial (g)-(i)Extension (j)-(l)Flexion }
   \label{exp_motions}
\end{figure}

\begin{figure}[hbt!]
    \centering \includegraphics[width=0.75\textwidth,height =2.5 in]{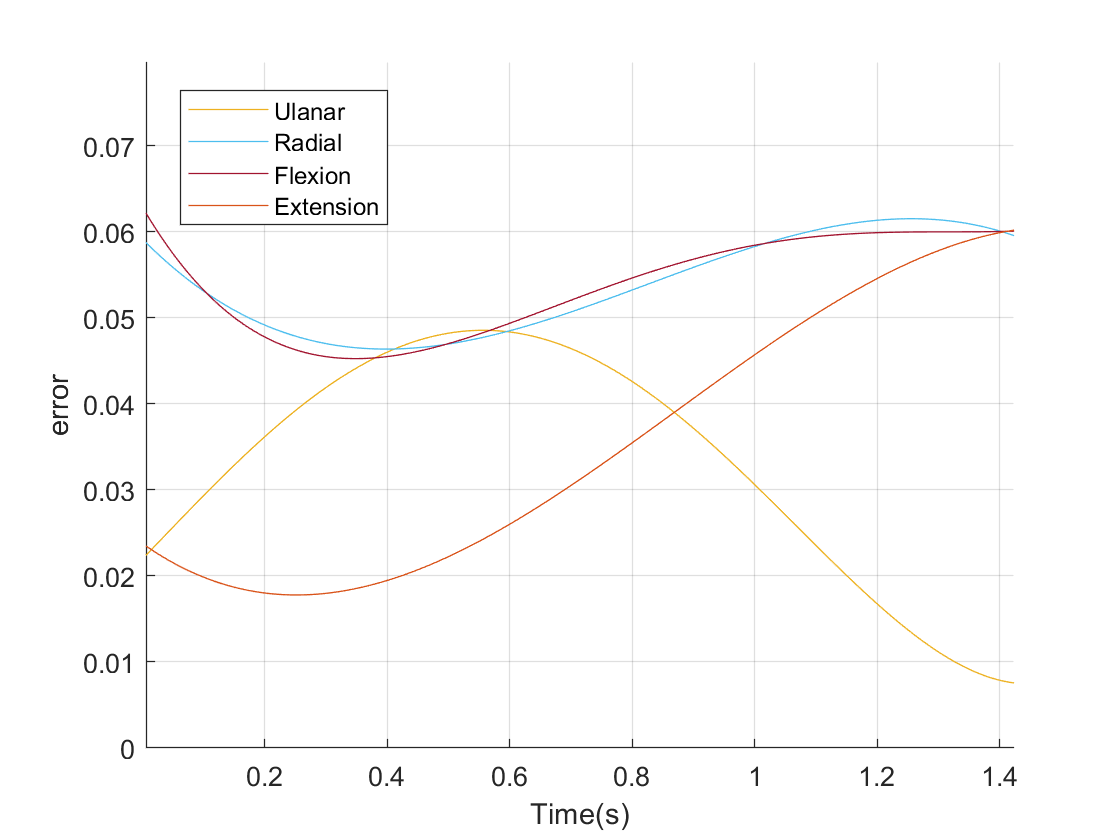}
    \caption{Error of motions during experimentations}
   \label{exp_traj}
\end{figure}

The experimental configuration for the developed wrist and hand model, along with its electronic components, is depicted in Fig. \ref{exp_setup}. An ArUco marker attached to the hand enabled the tracking of its poses during the experiments. The setup comprised four stepper motors, two motor drivers, a 3D depth camera, and an Arduino controller to facilitate real-time operations. Furthermore, ROS and MATLAB software were utilized for tracking the ArUco poses and implementing the control scheme, respectively. The hand's motion in all directions is illustrated in Fig. \ref{exp_motions} (a) - (l), with the associated motion errors presented in Fig. \ref{exp_traj}. Throughout the experimentation, the average RMSE values for deflection, settling time, and steady-state error across all directions as recorded as: $6\times 10^{-2}~ rad$, $1.35~s$, and $6\times 10^{-2}~rad$, respectively. The findings clearly indicated that the error values observed during the experimental phase are significantly higher than those recorded in the simulation study, primarily due to the lower stiffness of the springs used in the wrist segment.

\begin{figure}[hbt!]
    \centering \includegraphics[width=0.7\textwidth,height =2.5 in]{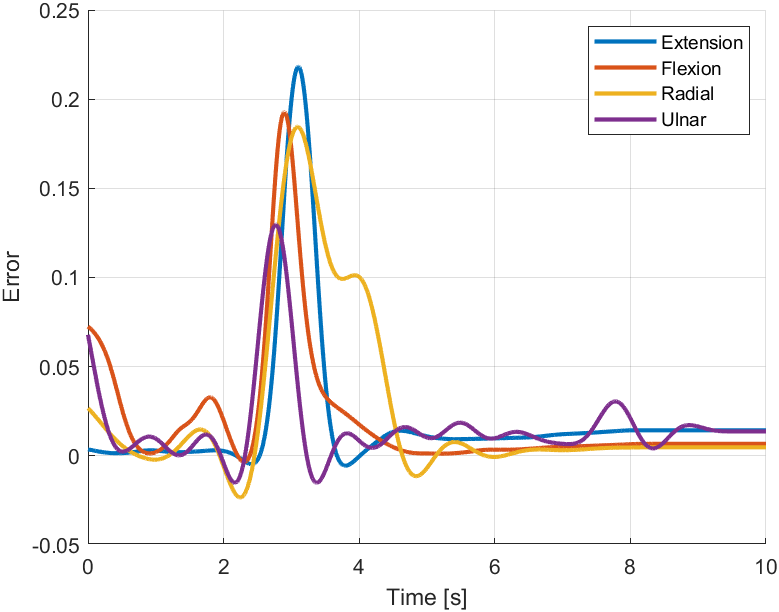}
    \caption{Error of motions in presence of force during experimentations}
   \label{exp_traj1}
\end{figure}

To evaluate the robustness of the proposed controller under external disturbances, an impulse force of $2\,\mathrm{N}$ was applied perpendicular to the wrist's motion at $t = 2\,\mathrm{s}$. As illustrated in Fig.~\ref{exp_traj1}, the tracking error increased immediately following the disturbance, reaching its peak shortly thereafter, and gradually diminished, stabilizing around $t = 5\,\mathrm{s}$ for extension and flexion directions. However, radial and ulnar motions converged around  6.7 s and 8.6 s respectively

\section{Conclusion}
An MPC system was developed utilizing a mathematical model derived from bending beam theory, specifically for the wrist section of a prosthetic hand. The proposed controller approach allowed the system to maintain the intended motion trajectories, effectively compensating for variations in the robot's physical properties and external environmental conditions. The integration of model improved controller efficiency by reducing computational time, while the MPC strategy contributed to a quicker response. Simulation results indicated that the proposed controller achieved a shorter settling time of less than  1.5 s, with the RMSE and steady-state errors remaining within acceptable limits. However, experimental RMSE values exceeded those from simulations, primarily due to inconsistencies in spring stiffness. Future research will focus on redesigning the wrist for greater structural integrity and enhancing controller strategies by integrating real-time sensor feedback to boost motion precision.

\section*{Acknowledgement}
This work was supported by the Italian Ministry of Research under the complementary actions to the NRRP ”Fit4MedRob - Fit for Medical Robotics” Grant (PNC0000007).


\begin{thebibliography}{25}

\bibitem{ref1}
Jyothish K.J., Mishra S.: A Survey on Robotic Prosthetics: Neuroprosthetics, Soft Actuators, and Control Strategies. ACM Computing Surveys 56(8), 1–44 (2024). doi: 10.1145/3648355

\bibitem{wmrac}
Gohari M., Sulaiman S., Schetter F., Ficuciello F.: A Sliding Mode Controller Design Based on Timoshenko Beam Theory Developed for a Prosthetic Hand Wrist. In: Proceedings of the 2025 11th International Conference on Automation, Robotics, and Applications (ICARA), pp. 338–342. IEEE, Italy (2025)

\bibitem{mpc1}
Spinelli F.A., Katzschmann R.K.: A Unified and Modular Model Predictive Control Framework for Soft Continuum Manipulators under Internal and External Constraints. In: Proceedings of the 2022 IEEE/RSJ International Conference on Intelligent Robots and Systems (IROS), pp. 9393–9400. IEEE, Japan (2022)

\bibitem{mpc2}
Johnson C.C., Quackenbush T., Sorensen T., Wingate D., Killpack M.D.: Using First Principles for Deep Learning and Model-Based Control of Soft Robots. Frontiers in Robotics and AI 8, 654398 (2021). doi: 10.3389/frobt.2021.654398

\bibitem{mpc3}
Pal A., He T., Wei W.: Sample-Efficient Model Predictive Control Design of Soft Robotics by Bayesian Optimization. arXiv preprint arXiv:2210.08780 (2022)

\bibitem{mpc4}
Yang Y., Zhang K., Chen Z., Li B.: Distributionally Robust Model Predictive Control for Constrained Robotic Manipulators Based on Neural Network Modeling. Applied Mathematics and Mechanics 45(12), 2183–2202 (2024)

\bibitem{mpc5}
Huang X., Rong Y., Gu G.: High-Precision Dynamic Control of Soft Robots with the Physics-Learning Hybrid Modeling Approach. IEEE/ASME Transactions on Mechatronics (2024). doi: 10.1109/TMECH.2024.3390169

\bibitem{mpc6}
Gonzales M., Polevoy A., Kobilarov M., Moore J.: Multi-Agent Feedback Motion Planning Using Probably Approximately Correct Nonlinear Model Predictive Control. arXiv preprint arXiv:2501.12234 (2025)

\bibitem{mpc7}
Kalibala A., Nada A.A., Ishii H., El-Hussieny H.: Real-Time Force/Position Control of Soft Growing Robots: A Data-Driven Model Predictive Approach. Nonlinear Engineering 14(1), 20250099 (2025). doi: 10.1515/nleng-2023-0121

\bibitem{mpc8}
Chen J., Zhao W., Meng Z., Mao D., Song R., Pan W., Zhang W.: Vision-Language Model Predictive Control for Manipulation Planning and Trajectory Generation. arXiv preprint arXiv:2504.05225 (2025)

\bibitem{ref15}
Sulaiman S., Menon M., Schetter F., Ficuciello F.: Design, Modelling, and Experimental Validation of a Soft Continuum Wrist Section Developed for a Prosthetic Hand. In: Proceedings of the 2024 IEEE/RSJ International Conference on Intelligent Robots and Systems (IROS), pp. 11347–11354. IEEE, Italy (2024)

\bibitem{prisma}
Liu H., Ferrentino P., Pirozzi S., Siciliano B., Ficuciello F.: The PRISMA Hand II: A Sensorized Robust Hand for Adaptive Grasp and In-Hand Manipulation. In: Proceedings of the International Symposium on Robotics Research (ISRR), pp. 971–986. Springer, Cham (2019). doi: 10.1007/978-3-030-95459-8\_60

\bibitem{bending}
Howell L.L.: Compliant Mechanism. McGraw-Hill, New York (2001)



\end{thebibliography}
\end{document}